\documentclass[sigconf, 	nonacm]{acmart}

\renewcommand\footnotetextcopyrightpermission[1]{} 

\usepackage{amsmath}
\DeclareMathOperator*{\argmax}{arg\,max}

\newcommand{\topic}[1]{\noindent \underline{ \bf #1}}
\usepackage[normalem]{ulem}
\usepackage{subcaption}
\useunder{\uline}{\ul}{}

\AtBeginDocument{%
  \providecommand\BibTeX{{%
    \normalfont B\kern-0.5em{\scshape i\kern-0.25em b}\kern-0.8em\TeX}}}

\begin{document}

\title{Semantic Annotation for Tabular Data}

\author{Udayan Khurana}
\email{ukhurana@us.ibm.com}
\affiliation{%
  \institution{IBM Research AI}
}

\author{Sainyam Galhotra}
\email{sainyam@cs.umass.edu}
\affiliation{%
  \institution{University of Massachusetts, Amherst}
}


\begin{abstract}
  Detecting semantic concept of columns in tabular data is of particular interest to many applications ranging from data integration, cleaning, search to feature engineering and model building in machine learning. Recently, several works have proposed supervised learning-based or heuristic pattern-based approaches to semantic type annotation. Both have shortcomings that prevent them from generalizing over a large number of concepts or examples.
Many neural network based methods also present scalability issues. 
 Additionally, none of the known methods works well for numerical data.  
  We propose $C^2$, a column to concept mapper that is based on a maximum likelihood estimation approach through ensembles. It is able to effectively utilize vast amounts of, albeit somewhat noisy, openly available table corpora in addition to two popular knowledge graphs to perform effective and efficient concept prediction for structured data. 
  We demonstrate the effectiveness of $C^2$ over available techniques on 9 datasets, the most comprehensive comparison on this topic so far. 
\end{abstract}

\maketitle

\section{Introduction}

Semantic annotation of structured data is crucial for  applications ranging from information retreival and data preparation to training classifiers. Specifically, schema matching component of data integration requires accurate identification of column types for the input tables~\cite{rahm2001survey}, while automated techniques for data cleaning and transformation use semantic types to construct validation rules~\cite{kandel2011wrangler}. Dataset discovery~\cite{zhang2020finding} and feature procurement in machine learning~\cite{kafe} rely on semantic similarity of the different entities across a variety of tables. Many commercial systems like Google Data Studio~\cite{googlestudio}, Microsoft Power BI~\cite{microsoftbi}, Tableau~\cite{tableau}  use such annotations  to understand input data, identify discrepancies and generate vizualizations.

Semantic annotation of a column in a table refers to the task of identifying the real-world concepts that capture the semantics of the data. For example, a column containing `USA', `UK', `China' is of the type \texttt{Country}. 
Even though semantic annotation of columns (and tables in general) is crucial for various data science applications, 
majority of the commercial systems use regular expression or rule based techniques to identify the column type. Such techniques require pre-defined patterns to identify column types, are not robust to noisy datasets and do not generalize beyond input patterns.

There has been recent interest in leveraging deep learning based techniques to detect semantic types. These techniques demonstrate robustness to noise in data and superiority over rule based systems. Prior techniques can be categorized into two types based on the type of data used for training and type of concepts identified. (a) Knowledge graph: Colnet~\cite{chen2019colnet} and HNN~\cite{chen2019learning} are the most recent techniques that leverage curated semantic types from knowledge graphs like DBPedia to generate candidate types and train classifiers to calculate the likelihood of each candidate. These techniques identify semantic types  for columns that are partially present in knowledge graphs and do not generalize well to generic types like \texttt{Person name}.
(b) Data lake:  Sherlock~\cite{sherlock} and Sato~\cite{sato} are the two recent techniques that consider concept labelling task as multi-class classification problem. These techniques leverage open data to train classifiers, however their systems are limited to the concepts which are also an exact match in DBPedia concept list. 

\begin{figure}[h]
    \centering
    \includegraphics[width=0.45\textwidth]{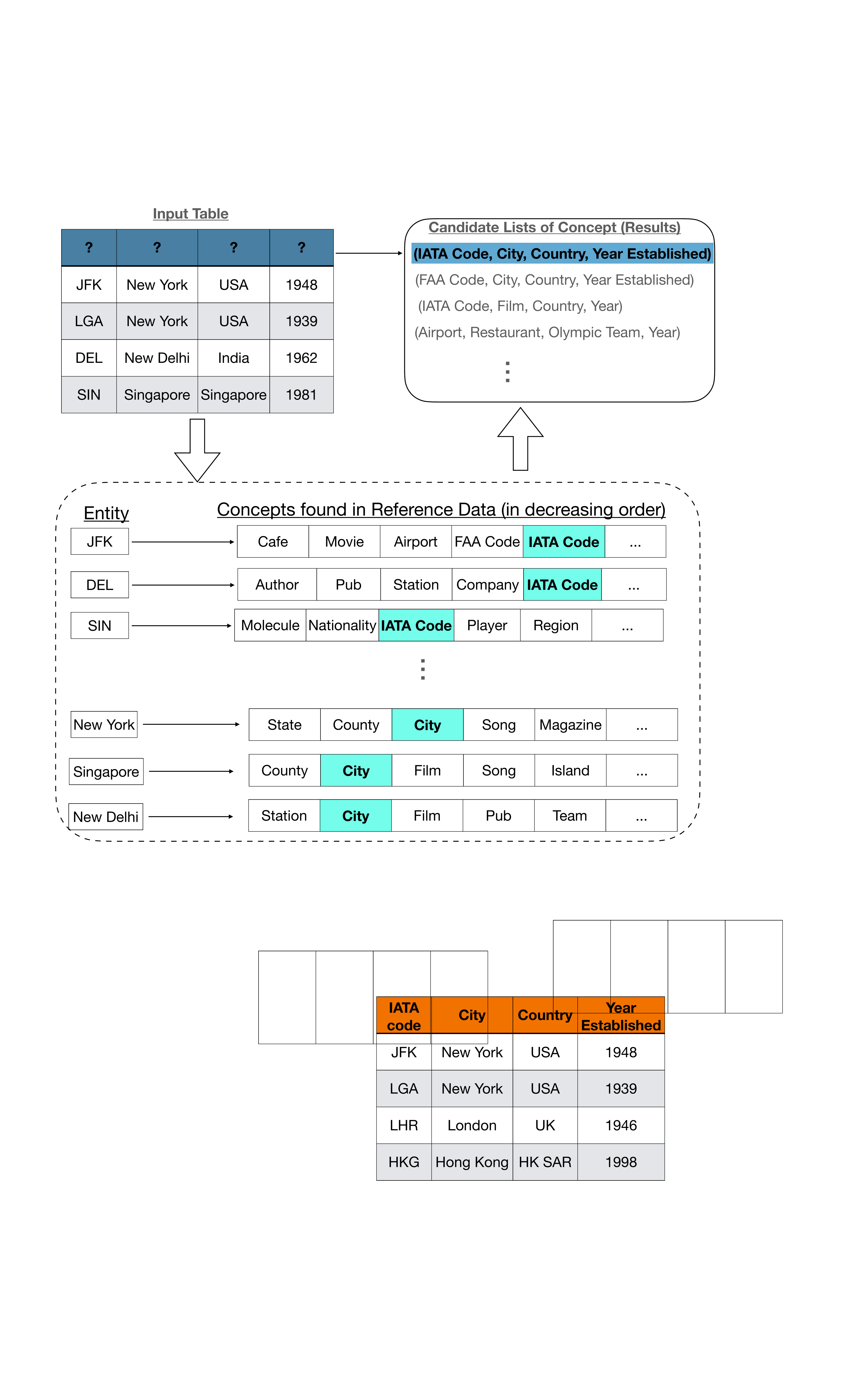}
    \caption{Example column to concept mapping using Maximum Likelihood Estimation.}
    \label{fig:airport}
   
\end{figure}

Even though these techniques have shown a leap of improvement over prior rule based strategies, they suffer from the following challenges. 
(a) Data requirement: Training deep learning based classifiers require large amounts of gold standard training data, which is often scarce for less popular types. There is a long tail of semantic types over the web; 
(b) Scale: Training deep learning networks for multi-class classification with thousands of classes is highly inefficient. Techniques like Colnet~\cite{chen2019colnet} train a separate classifier for each type and do not scale beyond $~100$ semantic types. 
(c) Numerical data: These techniques are primarily designed for textual categorical columns, i.e., ones with named entities, and do not work for numerical types like population, area, etc;  
(d) Open data: Colnet and HNN rely on the presence of cell values in curated knowledge graphs like DBPedia and do not leverage millions of web tables present over the web. On the other hand, Sato and Sherlock are able to utilize a small subset of open data lakes to train but their approach is forced to discard most of it, and they do not benefit from well curated sources like DBPedia, Wikidata, etc. either.
(e) Table context: Sato and HNN consider the table context, i.e., neighboring columns in predicting is considered through an extension of the above method using convolutional neural network style networks, where the problem of data sparsity intensifies; 
(f) Quality with scale: Besides the limited current ability, it is not clear how the approach of multiple binary classifiers or multi-class classifiers can at least maintain quality with the number of classes in 1000s. It is also not clear, how they would be able to utilize the nature of overlapping concepts - such as \texttt{Person} and \texttt{Sports Person}. 
\begin{example}
Consider an input table containing information about airports from around the world in Figure~\ref{fig:airport}. For the first column, \texttt{IATA code}, (the international airport code) is the most appropriate semantic concept. When searching structured data sources, it might not be the top choice for any entity (\textit{JFK}, \textit{SIN}, $\dots$), yet the challenge is to find it as the top answer. One may observe that while not the top choice in individual results, it is the most pervasive option. 
\end{example}

To address these limitations, we make the following observations.
(a) There are a plethora of openly available structured data from a diverse set of sources such as \texttt{\url{data.gov}}, Wikipedia tables, collection of crawled Webtables, and others (often referred to as data lakes), and knowledge graphs like DBPedia and Wikidata;
(b) It is true that all sources are not well curated, and one should consider a degree of noise within each source. However, a robust ensemble over the multiple input sources can help with noise elimination. While a strict classification modeling method may require gold standard data, a carefully designed likelihood estimation method may be more noise tolerant, and scale to larger contents of data; 
(c) Numerical data needs special handling compared to categorical entity data. While a numerical value is less unique than a named-entity (e.g., $70$ can mean several things), a group of numbers representing a certain concept follow particular patterns. Using meta-features like range and distribution help with quick identification of numerical concepts and are robust to small amounts of noise; 
(d) Instead of considering individual columns in isolation, global context of the dataset can help to jointly estimate the likelihood of each concept.

Our contributions in this paper are as follows: (1) Considering the above observations, we present $C^2$, a column to concept mapper using joint estimation through piecewise maximum likelihood estimation, utilizing large openly available structured data. (2) We provide special estimation methods for categorical, numeric, alphanumeric-symbolic data, respectively, while unifying them under a common framework of likelihood estimation; (3) We present novel and extensive indexes to support quick estimation computations for numeric, categorical and mixed type data; (4) We effectively perform context utilization in tables without polynomial increase in runtime or resources. (5) We provide the most comprehensive experiments on this topic using 9 datasets. We compare against 5 recent benchmarks and show our method to vastly outperform the benchmarks, in addition to being considerably faster.


\section{Related Work}
In this section, we discuss prior technqiues that have been developed to identify the type of a column.

\topic{Regular expression and lookup based techniques: } Use of manually defined patterns to identify the type of a column forms one of the components of various data science pipelines~\cite{trifacta}. Some of the recent applications include feature enrichment~\cite{kafe2}, schema mapping, data cleaning and transformation~\cite{jin2020auto,kandel2011wrangler}, structured data search~\cite{s3d} and other information retreival tasks. 

Many prior techniques have developed heuristics to identify such patterns. In order to reduce the  manual effort of enumerating patterns and rules, these techniques~\cite{kafe,oliveira2019adog,jimenez2020semtab,nguyen2019mtab,nguyen2019mtab,cremaschi2019mantistable,morikawa2019semantic,ritze2015matching} perform fuzzy lookup of each cell value over knowledge graphs to identify concepts. They assume that the cell values are present in knowledge graphs and are not robust to noise in values.

Deng et al.~\cite{deng2013scalable} presented a method based on fuzzy matching between entities for a concept and the cell values of a column, and ranking using certain simiilarity scores. Their technique is highly scalable but lacks robustness, as many movies have the same name as novel and a majority counting based approach would confuse a column of movies with that of novels. Other techniques that consider such heuristics include ~\cite{venetis2011recovering,kafe,oliveira2019adog,jimenez2020semtab,nguyen2019mtab,nguyen2019mtab,cremaschi2019mantistable,morikawa2019semantic, kafe2, s3d}.

Neumaier et al.\cite{neumaier2016multi} is specifically designed for numerical columns. Their approach clusters the values in a column and use nearest neighbor search to identify the most likely concept. It does not leverage column meta-data and context of co-occuring columns.

\topic{Graphical Models: } Some of the advanced concept identification techniques generate features for each input column and use probabilistic graphical models to predict the label. 

Limaye et al.~\cite{limaye2010annotating} use a graphical model to collectively determine cell annotations, column annotations and the binary column relationships. These techniques have been observed to be sensitive to noisy values and do not capture semantically similar values (which have been successfully captured by recent word embedding based techniques).

\topic{Learning approaches using Neural Networks: }
Chen et al.~\cite{chen2019colnet} present ColNet, a CNN-based approach for classification. It constructs positive and negative examples by looking up the cell values over DBPedia. These examples are then embedded using word2Vec to train CNN. The CNN is useful to build context amongst different cells of the column.
Chen et al.~\cite{chen2019learning} further extend prior methods to leverage inter-column semantics by training a hybrid neural network (HNN). This technique is extremely slow and requires more than 30 hrs to train over $40$ semantic types.
\cite{chen2019colnet,chen2019learning} rely on the knowledge graph lookup to generate candidates and do not extend to numerical attributes.

{Sherlock}~\cite{sherlock} is one of the recent techniques that models concept identification as a multi-class classification problem. It trains a multiinput feed forward deep neural network over a corpus of open data containing more than $600K$ columns referring to $78$ semantic types. All types with fewer than $1000$ training examples are ignored. The dependence of sherlock on large amounts of training data limits its applicability to the long tail of concepts.

SATO~\cite{sato} builds upon Sherlock by using context from co-occuring columns and jointly predicting the concept of all columns in the dataset. They use an LDA model, treating the table as a document to generate a vector of terms representative of the table context.  The inter-column context, however, is restricted to neighboring columns only, which may not work well in situations when the column ordering is irrelevant.

 \begin{figure}[h]
    \centering
    \includegraphics[width=0.5\textwidth]{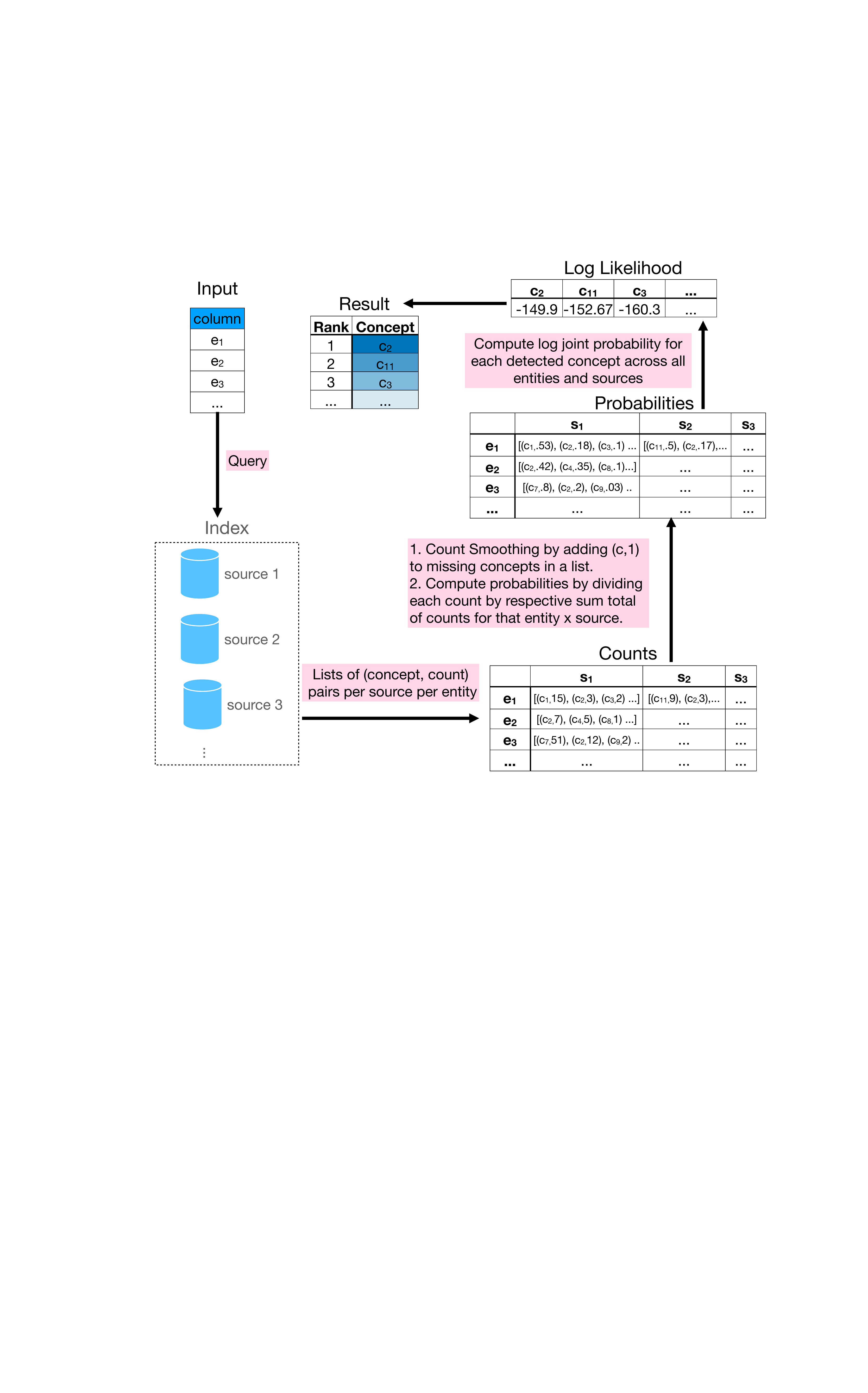}
    \caption{Concept prediction process illustrated for a hypothetical categorical entity column using our index, followed by smoothing, and maximum likelihood estimation. }
    \label{fig:column_process}
\end{figure}

\begin{figure}[h]
    \centering
    \includegraphics[width=0.5\textwidth]{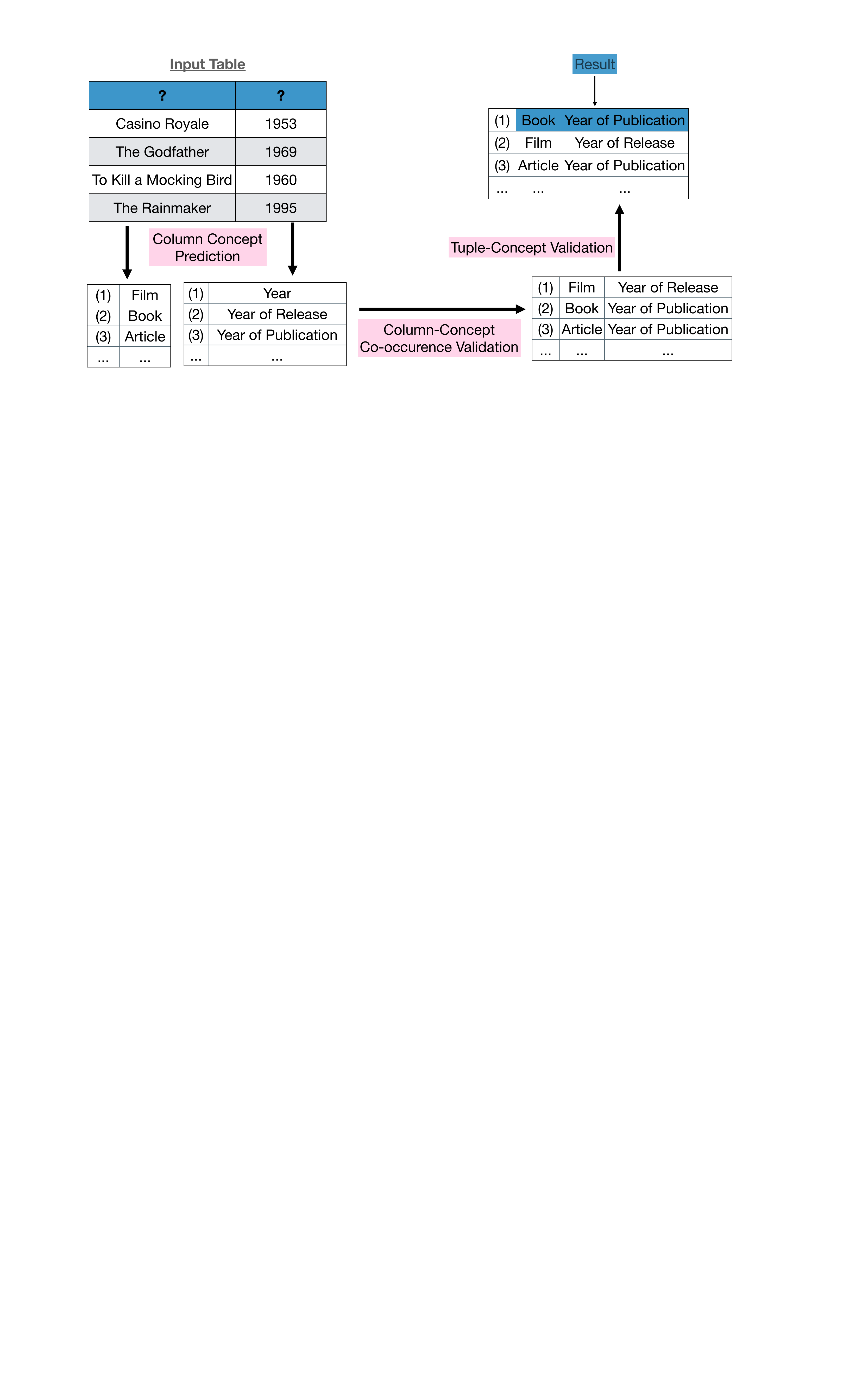}
    \caption{An example of effective tuple validation. All entities in first column are \texttt{books} as well as \texttt{movies}, the latter being more popularly mentioned across all sources. However, the second column's (year) values suggest that it is the year of book's publication, instead of movie's release.}
    \label{fig:book}
   
\end{figure}

\section{Preliminaries}
In this section, we discuss the setup and later define the problem.
\begin{itemize}
\item Given: a table $\mathcal T$ containing $m$ column vectors, each of which contains $n$ values. Alternatively, $\mathcal T$, also written as $\mathcal T_{nxm}$ may also be described as an ordered\footnote{the order in the set is only to preserve correspondence with the solution, an ordered list of concepts} 
list of the $m$ column vectors, i.e., $\mathcal T \{ v_0, v_1 \dots v_{m-1}\}$. Each vector contains $n$ values, $v_i = \{v_{i,0}, v_{i,1} \dots v_{i, n-1} \}$, for $0 \le i < n $. 

\item Let $\mathcal C$ be the set of all concepts in the English language. For practical purposes, this is the union of all property labels in DBPedia and Wikidata, all classes in DBPedia, and all frequently occurring (100 or more) column names in our table corpus.

\item $\omega(x) = c$ denotes that $c$ is the concept for entity $x$. Similarly, $\omega(v_i)$ denotes the concept for the column $v_i$. We treat concepts as a probabilistic notion. Concepts are considered neither exclusive nor absolute. They depend on context surrounding the object of question and exist in a degree of certainty or uncertainty, instead of definitiveness. This premise is crucial to the reasoning in our framework.

\item $\mathcal D$ is the data lake available for our reference. It is the reference against which we aim to optimize our estimation. As a simple case, if (where ever available), the {\em header} or{\em  title} of a column containing the value ``Chattanooga'' is \texttt{City}, 9 out of 10 times, and \texttt{football team}, in 1 out of 10 times, $Pr(\omega(chattanoonga)=city)=0.9$ and $Pr(\omega(chattanoonga)=city)=0.1$. For this study, $\mathcal D$ is a set of over 32 million tables originating from various sources~\cite{viznet, bhagavatula2013methods} and two knowledge graphs, Wikidata and DBPedia. In case of an entity from knowledge graph, its concept is found from an incoming {\em property label} or its class ({\em typeOf} property's value).

\end{itemize}



 \noindent \underline{\bf  PROBLEM:} For a given $\mathcal T_{mxn}$, find the set of $m$ concepts $\{ c_0, c_1 \dots c_{m-1} \}$, that, with the reference of $\mathcal D$ {\em \bf maximizes the likelihood} of:
\begin{itemize}
\item $\omega(v_{i,j}) = c_i$ for all $0 \le i < m$ and $0 \le j < n$ [likeliest concept for all cells of a column]
\item $c_i$ implying $c_k$ $\forall 0 \le i,k < m$, $i \ne k$ [likely co-occurence of a pair of concepts in the same table] 
\item $\omega(v_{i,j}) = c_i$ implying $\omega(v_{k,j})$  $\forall 0 \le i, k < m$, and $0  \le j < n $ [likely existence of these tuples or pairs of values belonging to said concepts]
\end{itemize}

It is important to note that our goal is not to find a solution that holds true for the above conditions; instead it is to find a solution -- an ordered set of concepts, that is the {\bf most likely} amongst all solutions possible as per the knowledge existing in $\mathcal D$.

\section{Approach}
In the previous section, we formulated the problem of concept mapping into three joint questions of likelihood maximization. 
However, it is a complex and expensive problem to solve. In order to achieve this objective, we need to evaluate an exponential number of candidates to find the ones that maximize the specified joint likelihood. Instead, we propose an alternative practical approach in which we solve a series of connected smaller likelihood optimization problems with certain assumptions (primarily thresholds) based on our observations, wherever suitable and necessary. We start with independently finding the top candidates for categorical entity columns. Using those as pivots, we narrow the search for numerical and mixed-type data columns (with the argument that a column like \texttt{Population} is likely to be found in a table talking about a \texttt{Country}, or \texttt{City}, as opposed to a table contianing \texttt{Movie}). Considering ranked (top-k) candidate concepts for all columns, we finally reason which narrative, i.e., the combination of concepts, best explains the given table with respect to the millions of records in our reference data. In this section, we explain our proposed estimation methods and in the next section, we describe the necessary data processing and data structures required to efficiently realize this logic.

For an $n$-valued column vector,  $v_i = \{v_{i,0}, v_{i,1} \dots v_{i, n} \}$, let $\Pr(v_{i,j} = c_l))$ denote the probability that the entity $v_{i,j}$ belongs to the concept $c_l$. Now, $\sum_{l \in \mathcal C}  \Pr(v_{i,j} = c_l)) = 1$, i.e., all probabilities of concepts representing an entity sum up to 1. We estimate these probabilities from our corpora $\mathcal D$. The process is described in section~\ref{sec:details} and illustrated in Figure~\ref{fig:column_process}. For now, it is sufficient to keep the ``\textit{Chattanooga}'' example in mind.

For each entity $v_{i,j}$, there exists a list of probabilities $\Pr(v_{i,j} = c_l)),\forall c_l\in \mathcal{C}$. And each entity in a column presents different probabilities for different concepts. We present that:

\begin{itemize}
\item The likelihood of all entities of a column belonging to concept $c$ is the joint probability of all its entities belonging to the concept $c$ at the same time, i.e.,  

$ L(\omega(v_i=c)) =  \Pr(\omega(v_{i}) = c)\\  =  \Pr(\omega(v_{i,0})=c)  \times \Pr(\omega(v_{i,1})=c)  \dots  \Pr(\omega(v_{i,n_1})=c)\\ = \prod \Pr(\omega(v_{i,j}) =c) $, assuming independence.

\item The concept of a column is the concept that most likely explains all of its components, i.e., $\omega(v_{i}) = MLE(c)$, $\forall c \in \mathcal C$.

$c* =  MLE(c) = \argmax\limits_{c \in \mathcal{C}}  L(\omega(v_i)=c))$

\item For computational convenience, we use Log likelihood,
$ c* =  \sum Log \Pr(\omega(v_{i,j}) =c) $
\end{itemize}

\topic{Multiple Sources of Data:} 
Multiple sources of background information provide a rich and diverse set of data and concepts, which is particularly useful to improve our predictions through an ensemble\cite{dietterich2000ensemble}. Instead of treating the entire reference data as one large source, and counting and calculating probabilities from it, we utilize the diversity provided by these sources to aid in likelihood estimation. If we were to collapse all sources into one, a concept that is heavily dominating in a particular source (with very high presence for an entity) and while insignificant in other sources, could overwhelm all other concept options. Diversity through ensembles rewards consistency for our benefit. Note that this diversity is in addition to the one introduced by results for various entities in the column, and the same notion of a concept's consistency through out the entities is rewarded through the proposed method. For, $|\mathcal D|$ sources and $n$ entities in the column, we obtain $n \times |\mathcal D|$ lists of concept-count arrays to perform likelihood estimation on. Figure~\ref{fig:column_process} illustrates this process.
  
\begin{itemize}
\item $ c* =  \sum Log ( \alpha_{d} \Pr(\omega_{d}(v_{i,j}) =c)) $, for all $d \in \mathcal D$ and $\forall 0 \le i < m$, and $0  \le j < n $
\item $\alpha_d$ is the weight assigned to each source based on a simple regression experiment. 
\end{itemize}

\topic{Smoothing: } Since each entity of the column fetches a different set of concepts from each source, and due to our multiplicative joint-probability estimation, we perform smoothing~\cite{simonoff2012smoothing} of the concept-count pairs. For each concept that occurs in the list fetched for any entity of a certain column, wherever that concept is not found, we will add it with the minimal count (1), so that the probability of its existence is non-zero, yet a small number. This is particularly useful in cases of unseen entity values, such a a new \texttt{video game}, and while the results corresponding to other entities strongly suggest that concept, we do not wish to lose it as a candidate just because one of few entities or sources do not contain that information.

\topic{Concept co-occurence: }
A table's columns usually represent related concepts. Sometimes a table is based around a central column, with other columns representing its attributes, such as the census data tables containing attributes of an individual -- from \texttt{name} (which can be thought of as the central concept, the \texttt{person}) and attributes like \texttt{age}, \texttt{income}, etc. Other tables may be less decentralized, especially {\em joinable tables}. In either case, column concepts are usually linked to each other directly or indirectly. So far in our concept inference, we have only considered a column independently, conducting maximum likelihood estimation with respect to only its values, but not considering the concepts or values of co-occuring columns.

To obtain appropriate column context, we prepared the information of all column-name pair co-occurrences in the reference data $\mathcal D$. This gives us the ability to reason how likely are two concepts to co-exist in a given table. We utilize this for the purpose of restricting the scope of search for numerical and the mixed-type data, as well as re-ranking the final combination of $m$ concepts as follows:

\begin{itemize}

\item We perform the concept search for categorical entity-based columns first, due to inherently less ambiguity than numerical attributes. Based on the results (top-k answers) of categorical entity attributes, we influence the search performed by \texttt{Numerical Column Detection}. We search the up to 25 possibly co-occuring concepts for each numerical column. While the mixed-type columns have less ambiguity than numerical columns, we search for their concepts after the categorical entity columns as well, as these attributes are rarely the central concept in any table. 
\item If $c_1$ is the concept for a categorical column ($v_1$), Using Bayes' theorem, probability for a numerical column ($v_2$) also in the table to represent the concept $c_2$ is: $\Pr(\omega(v_2) = c_2 | \omega(v_1) = c_1)  = Pr(\omega(v_1) = c_1 \wedge \omega(v_2) = c_2 ) / Pr(\omega(v_1) = c_1)$.  
\item However, $v_1$ has other possible concepts with different probabilities. And there are other categorical columns with an array of probable concept values (with various probabilities). The complete probabilistic estimation of a numerical column's concept is a probabilistic average of the above. We omit the mathematical expression here.
\item Note that the above weighted conditional probability is used to determine the scope of search for \texttt{Numerical Column Detection}. Each candidate there is then evaluated based on the \texttt{Numerical Interval Tree} for that concept, as explained in Section~\ref{sec:details}.
\item Once we have obtained the candidate results for each column, we once again compute the joint probability for each candidate list $[ c_0, c_1 \dots c_{m-1} ]$, by approximating $\Pr( c_0 \wedge c_1 \dots c_{m-1})$.
\end{itemize}


Column co-occurence validation provides us with improvement in quality of predictions as well as efficiency by reducing the scope of numerical and mixed-valued candidates to consider. 

\topic{Tuple Validation:} Consider the case in Figure~\ref{fig:book}, where the first column appears to represent well-known movies, and the second column contains year. With help from column co-occurrence check, we find a popular pair of column (\texttt{movie}, \texttt{year released}). While we have considered the relationship between column concepts, upon closer examination it is revealed that the combination of title and year actually represent a book of the same name instead, along with year of publication. For instance, the book {\em The Godfather} was released in $1969$, whereas, the movie of the same name was released in 1972. Even though \texttt{movie} is a more popular concept in our database and occurs almost thrice as frequently as a book, as well as the combination of movie and year being frequent, the validation of tuples gives credence to the concept pair of  (\texttt{book}, \texttt{year published} to be more likely.

\begin{figure*}[h]
    \centering
    \includegraphics[width=\textwidth]{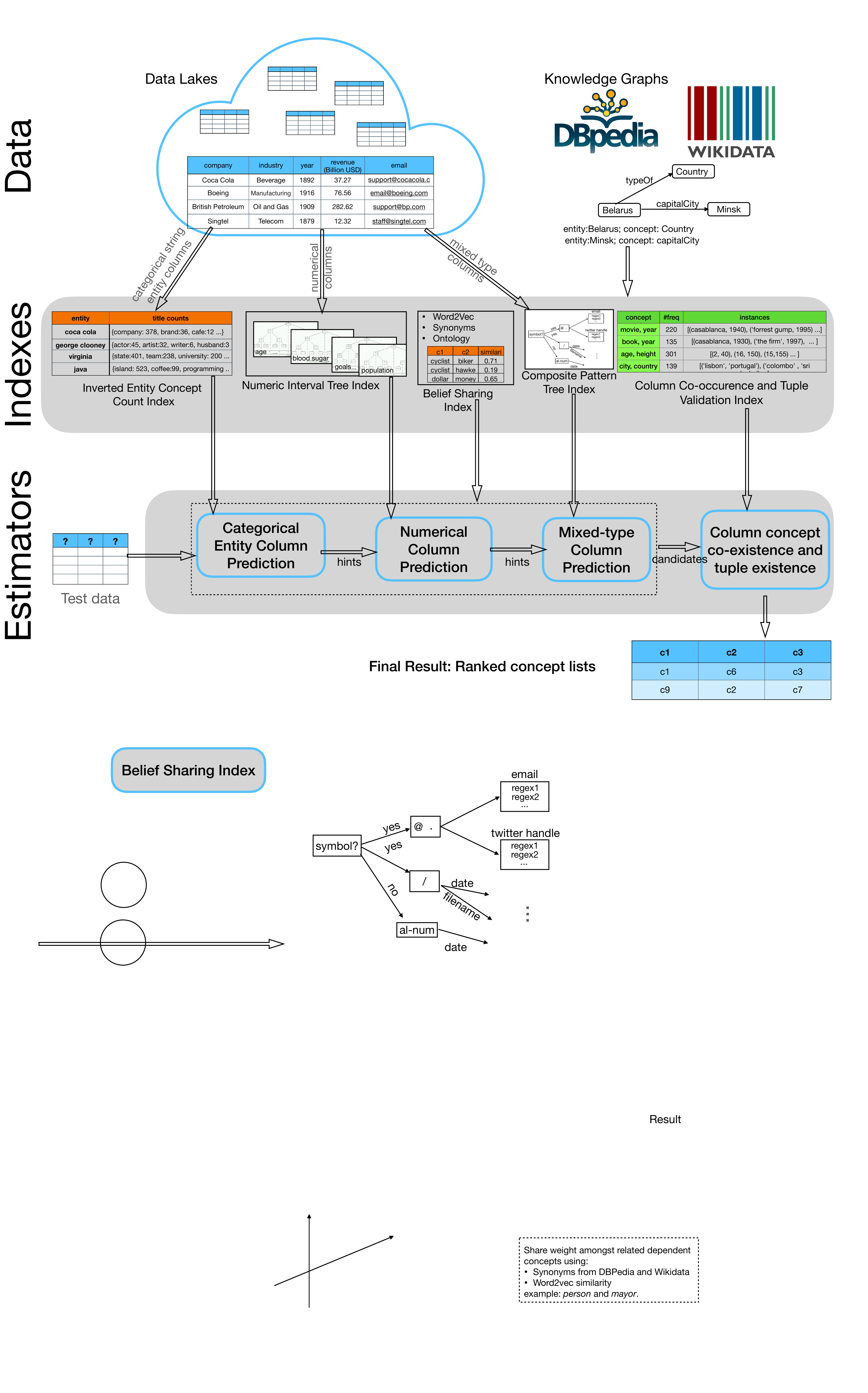}
    \caption{System Architecture depicting the layers of (a) {\em reference data}, i.e., knowledge graphs and web tables; (b) different {\em  indexes} based upon the column data type; and (c) the {\em estimators} that perform inference on a given table using the indexes.}
    \label{fig:overview}
\end{figure*}

We perform this operation by looking for tuple pairs in the \texttt{Column Co-occurrence and Tuple Validation} index (Section~\ref{sec:details} and Figure~\ref{fig:overview}) for up-to ten randomly selected rows. The prior likelihood of all candidates is multiplied by the number of occasions on which a candidate led to a match, unless a candidate did not match at all, in which case it is left unchanged. The matching for numerical attributes is performed within $0 \le \epsilon < 1$ multiplicative error (default 0.1) , where as the categorical entities must be exact. Each validated pair must contain at least one categorical entity. For example, in tuple 
\texttt{Mountain:} \textit{Mount Everest} and \texttt{Height:} \textit{8,884}, a match must have {\em Mount Everest} as a mountain and the height within $10\%$ of $8884m$. If the potential height column consisted a value of say $1$, then it was either an error or another concept.

\topic{Belief Sharing:}
Several column headers collected from different sources may refer to the same concept. For example, all \texttt{scientists} are \texttt{persons} and some persons are scientists. Any \texttt{person} can be considered a \texttt{human} and vice-versa. The reference data which originates from multiple sources has no standardized way of naming columns. When we collect counts from the index and subsequently compute probabilities, it considers concepts as independent. However, in reality, the related concepts are not independent. Hypothetically, if an entry was marked \texttt{Human} 30\% of the times, \texttt{Person} 30\% of the times, and \texttt{Object} 40\% of the times, the correct answer is either \texttt{Human} or \texttt{Person}, with 0.6 probability each, and not \texttt{Object}. The probabilities after {\em Belief Sharing} do not add up to one, and the concepts aren't considered independent of each other anymore. We perform this belief sharing through (a) Word2Vec~\cite{mikolov2013distributed} similarity; (b) and a list of synonyms extracted from DBPedia and Wikidata; (c) concept hierarchy from DBPedia and Wikidata. 
If two concepts are found to be synonyms, their counts are merged. Else, if there is a parent-child relationship found in the hierarchy, then a parent transfers equivalent of its count (while retaining its own), to its children in the proportion of children's existing counts. Whereas, a child transfers half of its count  to its parent (while retaining its own count). Finally, if none of those two criteria are matched, we resort to Word2Vec, and if a pair of concepts is found to be at least 40\%\footnote{The $0.4$ similarity threshold for similarity was determined based on empirical search.} similar, then they transfer their respective counts to each other in the fraction of similarity (while retaining their own). This step has a profound impact on the quality of predictions, as it is able to provide a convergence between the diverse sources following different nomenclature and other concept naming practices. This also happens to be expensive, which is partially alleviated by indexing word similarity.


\section{Data Preparation}
\label{sec:details}
In this section, we discuss the reference data preparation, which is crucial to improve the runtime of  query processing in $C^2$. We describe our indexing components for categorical string data (inverted count index), numeric data (a modified interval tree), and mixed-type data (a new pattern search tree). The approach of such indexing for column type detection is novel and unique to $C^2$. For comparison, at runtime, Colnet fetches information from DBPedia based on each cell-values and hence takes hours to perform queries which $C^2$ performs similar lookup in less than $10$ seconds. Moreover, the other approaches do not have ways to address numerical column type detection\footnote{except selective cases such as \texttt{year}, which is practically a categorical attribute.} because their embedding methods are inadequate for numerical values.


\topic{Inverted Entity Concept Count Index: }
This key-value inverted index lists the number of times a categorical entity has been mentioned under respective column names. This is done separately for each source of data. For the knowledge graph triples, it the column name and entity are replaced by (a) a property and the object, respectively; or (b) entity, and object of its``typeOf'' property (indicating class name, if any), respectively. Its construction is described briefly as follows.  
For a categorical entity column in a table in $\mathcal D$, with column name $c$ and column values as $e_0 , e_1 \dots e_{l-1}$, a key-value pair $e_i : (c, 1)$ is produced $\forall 0 \le i < l$. All pairs are aggregated based on the key and the counts for each column name or concept are aggregated into a list. The final index contains the keys as the entity names, where the values are lists of pairs (concept, count). This is performed separately for each source of data and the entity keys are prefixed by a code for the data source. The utilization of this index is illustrated in Figure~\ref{fig:column_process}.

\topic{Numerical Interval Tree Index: } Numbers need special treatment when compared to categorical entities. For instance, the number \texttt{70}, could be the age of a person, systolic blood pressure value, number of wins by a team in a season, ad infinitum. Whereas, categorical entities like ``University of Virginia'' are less ambiguous, belonging to a much smaller possibility of candidate concepts. On the other hand, a bunch of numbers provide us with an additional piece of information, a linear scale to check whether a set of numbers belongs to a a range of numbers typically associated with a concept. So, while 70 might have possibly meant the age of a person, the group of values, \{ 70, 100, 130, 120, 140\} collectively are unlikely to represent the age of a person from our data. Hence, we need not check the exact values but the likely range of a concept against that of the given data. 

We design a modified {\em interval tree} for this purpose. For each numerical concept in the reference data, we build a separate tree where we effectively compare the ranges of the given set (an interval) with all historic range in which that concept has occurred. Through {\em interval queries}, we find the number of intervals that intersect with the given column's range, i.e., [min, max]. The modification with  the traditional interval tree is that instead of the leaf nodes linking to objects falling in the range of atomic intervals, our tree instead consists of the number of times such a concept was true if the range was that atomic interval. This operation works in $O(log n)$ in the number of intervals indexed in that tree, i.e., belonging to that concept in our data. For fundamental details on external interval trees, we refer the reader to Arge et al.~\cite{arge1996optimal}. 

Once this count is obtained, we divide it by the total number of intervals/ranges for concept $c$ in reference data. For range, $r$, this would provide us with the conditional probability, $Pr(r|c)$. However, we are actually looking for $\Pr(c|r)$. Now, if $\Pr(c)$ is the probability of occurrence of this concept (provided it is a numeric column, since that is our context of this index), which can be obtained by simply the ratio of occurrence of concept $c$ in all of numeric columns seen in the reference data. Using Bayes' theorem:

\indent \indent \indent \indent \indent $\Pr(c | r) = {\Pr(r|c) \times \Pr(c) \over \Pr(r)} \propto \Pr(r|c) \times \Pr(c) $

For all candidate concepts $c$, the denominator ($\Pr(r)$) is constant and can be ignored. Thus doing such search over all the numeric concept trees can tell us the most likely column concept, without any other table context available. 
Cost of construction of each tree is $O(n\cdot log n)$ and update cost is $O(log n)$.

\begin{figure}[h]
    \centering
    \includegraphics[width=0.4\textwidth]{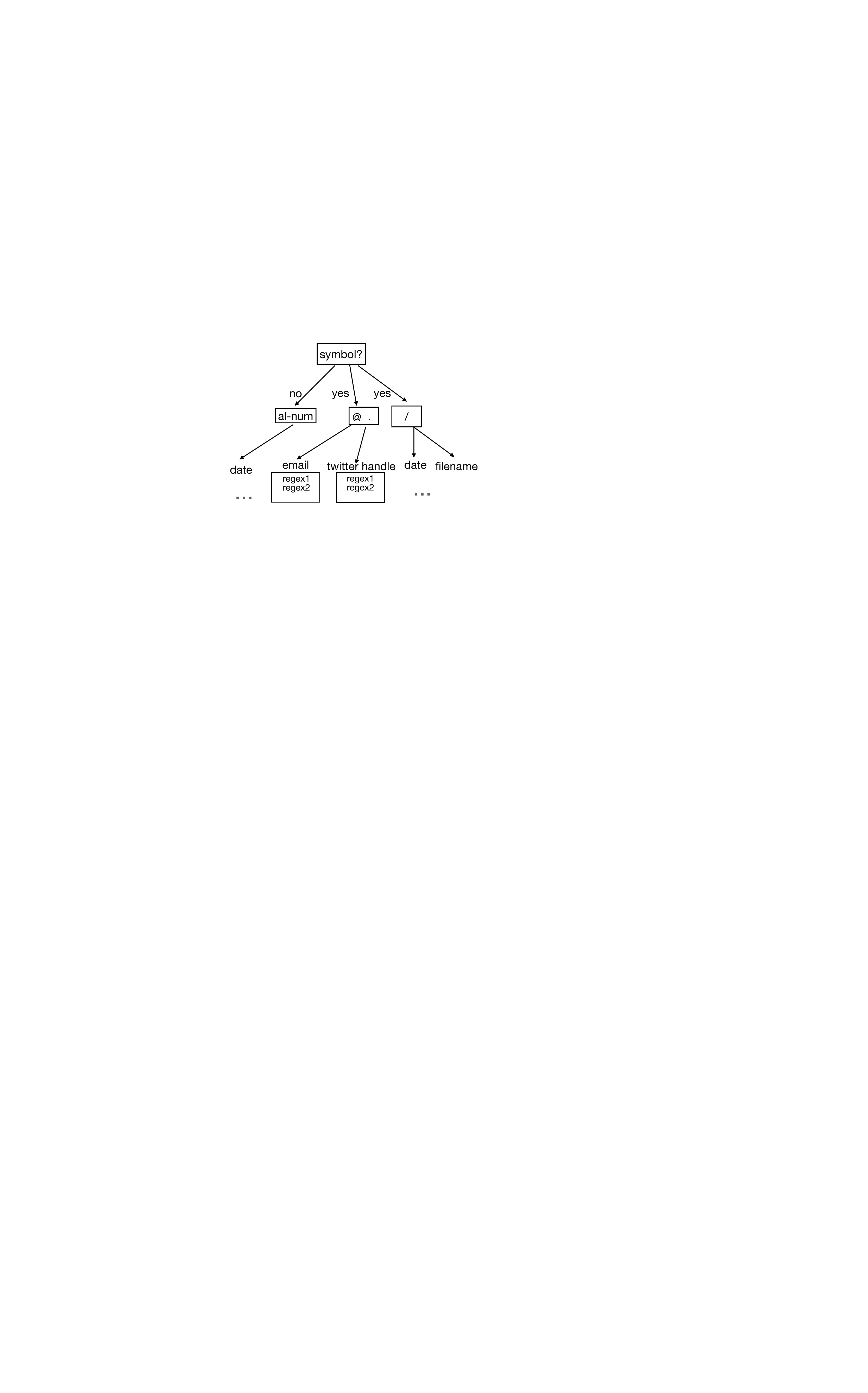}
    \caption{Composite Pattern Index for mixed-type data. }
    \label{fig:regex}
\end{figure}

\topic{Composite Pattern Tree Index:} This index deals with data that is neither categorical/string type nor the numeric type (and neither the natural language type, which is beyong the scope of this work). Instead, it deals with the mixed type of alpha-numeric-symbolic data such as emails, dates, website addresses, DOIs, computer filenames, and so on. We run a \textit{Regular Expression} (Regex) generator\footnote{TDDA: https://tdda.readthedocs.io/en/v1.0.30/rexpy.html}, on each such column in the reference data and aggregate and deduplicate the regexes by each concept or column name. The index is a decision tree, which separates the cases by whether an entity contains symbols or not, followed by if they contain one symbol or the other, and so on. The leaf nodes contain links to groups of regexes, all corresponding to one concept. For example, as seen in Figure~\ref{fig:regex}, the main separator for emails is whether the entity contains `@' and `.' symbols. One concept may occur in clusters at different leaf nodes. For example, \texttt{Date} of the format `11/07/2005'  and `November 07, 2005' land up in separate {\texttt Date} clusters. The clusters are found by analyzing all regexes and columns belonging to a concept, and for any symbol combination occurring above a threshold (10\%) of the total columns for that concept, a new leaf node is created. At runtime, when an input column is considered, it may well qualify for multiple branches of the tree. For example `@dan.singer' qualifies for both \texttt{Instagram handle} as well as \texttt{Email}, and regex-sets corresponding to both root nodes will be evaluated. The evaluation and inference follows the same logic as that of a concept's tree in Numerical Interval Tree Index described above. The number of regexes in a leaf node set are limited to 100 using random selection.

\topic{Column Co-occurence and Tuple Validation Index: } This key-value index stores the following: (a) concept-pairs in the reference data; (b) the frequency of that pair in the reference data; (c) Which entity-pairs exist under that concept pair and how frequently. This index has two important uses: (a) find the likelihood of two concepts co-existing; (b) find how likely is a pair of entities possible under a concept pair. A illustration of this index can be seen in Figure~\ref{fig:overview} as the rightmost index.

\topic{Belief Sharing Index: } We index concept pair similarity scores using a pre-trained Word2Vec model~\footnote{https://code.google.com/archive/p/word2vec/} as we compute them once. In addition, the synonyms and the concept hierarchy from DBPedia and Wikidata are read to an in-memory hash during initialization.


\section{Evaluation}
In this section, we empirically evaluate the effectiveness of $C^2$ on a collection of nine real-world datasets and compare it with prior literature.

\subsection{Datasets}
We evaluate results on a collection of 9 datasets, the most comprehensive so far. Out of these, 8 are from existing papers or competitions -- Limaye~\cite{limaye2010annotating}, Semantification~\cite{kacprzak2018making}, SemTab challenge rounds 1 to 4~\cite{jimenez2020semtab}, T2Dv2~\footnote{T2Dv2: http://webdatacommons.org/webtables/goldstandardV2.html}\cite{lehmberg2016large}, ISWC17~\cite{efthymiou2017matching}. Additionally, we generated the 9th dataset by manually tagging a subset of columns from a 10\% random sample of the Manyeyes data~\cite{viznet}; all these tables were removed from our reference data. This dataset contains categorical, numerical and mixed types. Semantification contains numerical and categorical. Rest all contain categorical type columns only.

\begin{table}[]
\small
\begin{tabular}{|l|l|l|l|l|l|l|}
\hline
{ Code}   & { Name}      & { Cases} & {Concepts} & { Cat} & { Num} & { Mixed} \\ \hline \hline
\textbf{LIM} & Limaye          & 84          & 18             & No                & No              & No          \\ \hline
\textbf{SEM} & Semantification & 492         & 140            & Yes               & Yes             & No          \\ \hline
\textbf{ST1} & SemTab Round 1  & 120         & 32             & Yes               & No              & No          \\ \hline
\textbf{ST2} & SemTab Round 2  & 14780       & 356            & Yes               & No              & No          \\ \hline
\textbf{ST3} & SemTab Round 3  & 5762        & 266            & Yes               & No              & No          \\ \hline
\textbf{ST4} & SemTab Round 4  & 1732        & 118            & Yes               & No              & No          \\ \hline
\textbf{T2D} & T2Dv2           & 411         & 55             & Yes               & No              & No          \\ \hline
\textbf{ISW} & ISWC 2017       & 620         & 31             & Yes               & No              & No          \\ \hline
\textbf{OUR} & Ours            & 120         & 56             & Yes               & Yes             & Yes         \\ \hline
\end{tabular}

\caption{Characteristics of datasets used for evaluation, Code name used in plots, datasets name, number of test cases, number of concept classes, if it contain categorical attributes (Cat), numerical attributes (num), mixed attributes (mixed). }
\end{table}

\subsection{Setup}
All codes were implemented in Python and run on a server with 128GB RAM. 
We observed  following challenges in evaluating prior concept mapping systems. First, due to the lack of standardized concept ontology\footnote{DBPedia ontology is well-defined but quite small.} in majority of the datasets, there is a lot of  discrepancy in the target names representing the same concept such as \texttt{Film} and \texttt{Movie} are common and an issue with evaluation. Secondly, presence of overlapping concepts such as \texttt{Region} and \texttt{Province} presents another challenge. If the test case expects one but a system provides the other, accepting it as the correct answer overlooks the fine semantic distinction; however, rejecting this match takes away the partial credit deserved by the system. A special case of this is the parent-child relationship. For example, if the precise answer is \texttt{Mayor}, should \texttt{Person} be accepted as a correct answer? While all the mayors are persons, such a classification falls short of the desired precision.

We have purposefully stayed away from partial scoring (between 0 and 1) in above cases due to the lack of an appropriate framework. For a fair comparison, we manually investigated the output of various techniques and collected a ``reasonable'' set of synonyms in the English language for all concepts in the ground truth, such as \texttt{Film}: \texttt{Movie}, \texttt{Region} and \texttt{Administrative Region}, etc. Ontology based substitutions for parent-child relationships, where the parent is a broad category, such as \texttt{Place}, \texttt{Thing}, \texttt{Agent}, were not considered. However, we found two datasets from the ISWC SemTab challenge track which have done that explicitly using available hierarchies (datasets ST2 and ST3) and we considered those in our evaluation.

\topic{Baselines.} We compared $C^2$ to three recent openly available recent systems -- Sherlock~\cite{sherlock}, SATO~\cite{sato}, Colnet~\cite{chen2019colnet}, to the extent we could run their code in a reasonable limit of time (48 hours per dataset). In addition, we provide experimentation providing insights into the components of our system.
We compute  accuracy of the top-ranked concept to compare the quality of returned concepts and time taken to predict the concept for efficiency.

\subsection{$C^2$ Performance}
\begin{figure}
    \centering
\includegraphics[width=\columnwidth]{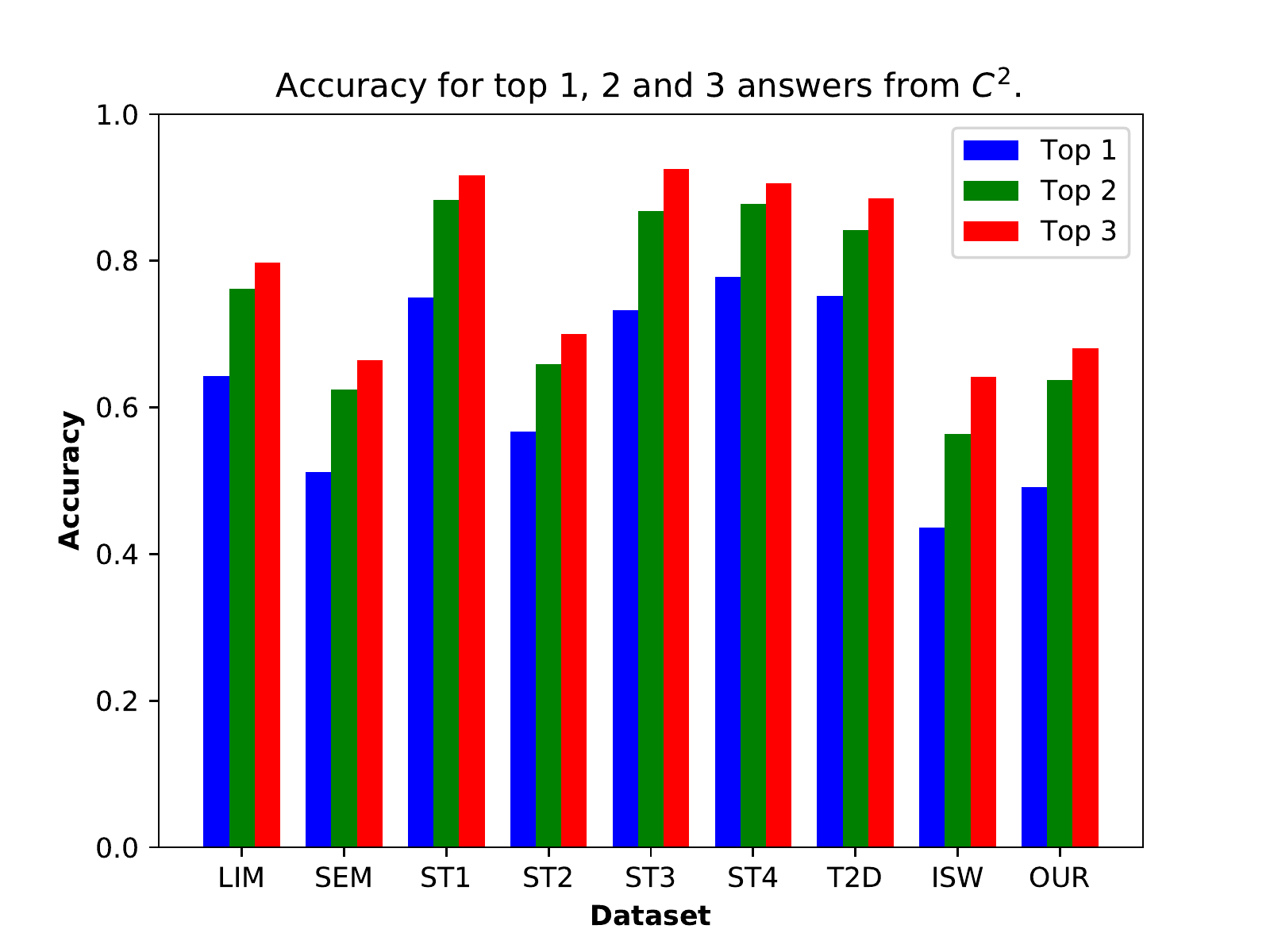}
    \caption{Accuracy of $C^2$'s top 1,2, and 3 (cumulative) results across different datasets.}
     \label{fig:C2C_123}
\end{figure}

Figure~\ref{fig:C2C_123} compares the accuracy of top 1,2 and 3 concepts returned by $C^2$ on the considered datasets. 
Across all datasets, the top-3 candidates identified by $C^2$ contain the correct concept with more than $80\%$ accuracy for $5$ out of $9$ datasets, and an average of $62.9\%$ for the top result, $74.6\%$ for either of top-2 results and $79.1\%$ for either of top-3. $C^2$ performance on numerical attributes alone (in SEM and OUR dataset), is $41.2\%$, $44.3\%$ and $45.1\%$ on top 1,2 and 3 results respectively.


\topic{Impact of Belief Sharing:} When $C^2$ is run with \texttt{Belief Sharing} turned off, it's average accuracy drops to $51\%$, $64.2\%$ and $75.4\%$ for top 1,2  and 3 results, respectively.

\topic{Impact of Concept Pair and Tuple Validation:} When $C^2$ is run with \texttt{Concept Pair and Tuple Validation} turned off, it's accuracy drops to $60.6\%$, $71.8\%$ and $75.9\%$ for top 1,2  and 3 results, respectively. 

\subsection{Comparison with Sherlock and SATO}
\begin{figure}
    \centering
\includegraphics[width=\columnwidth]{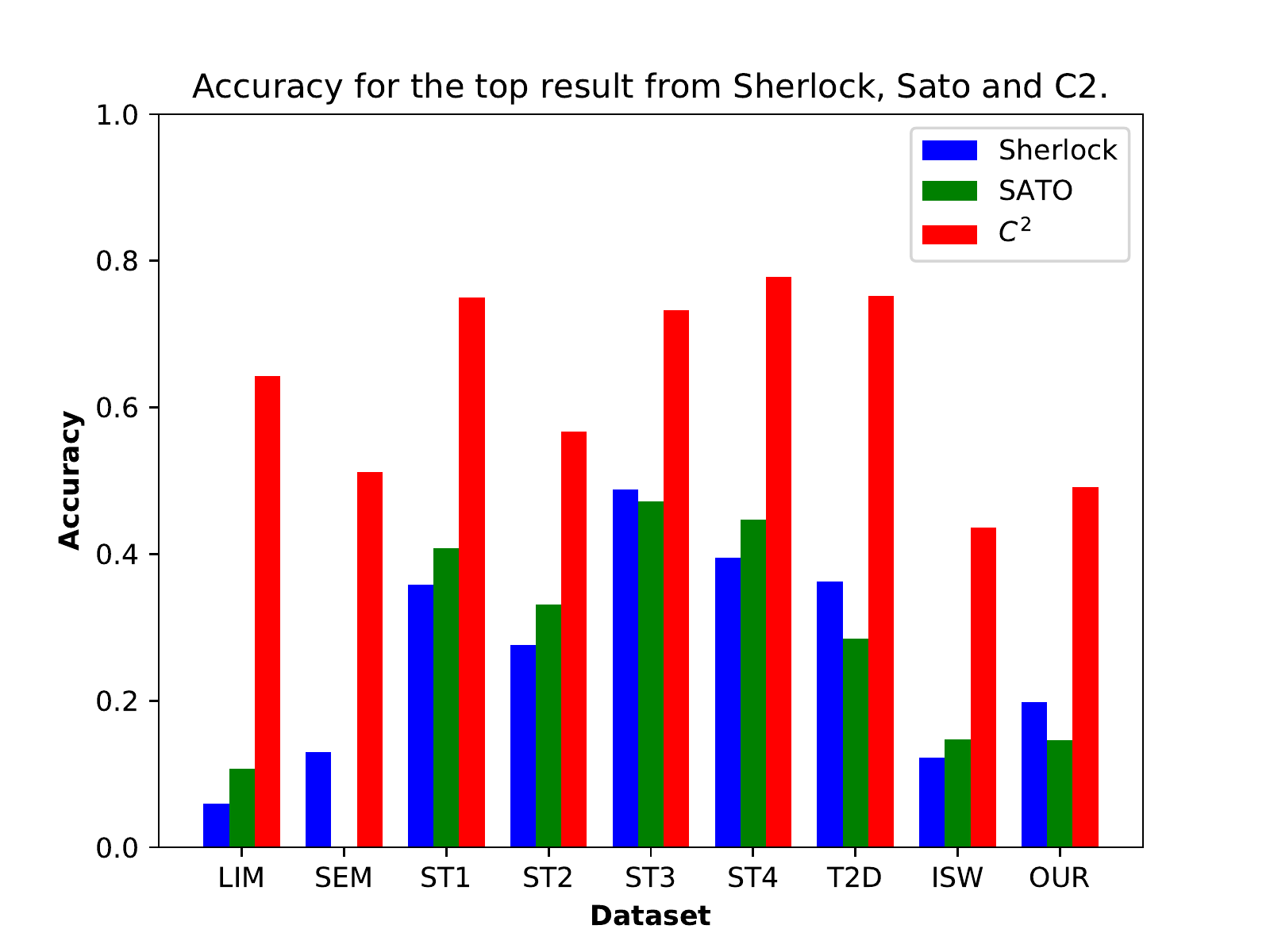}
    \caption{Comparing Sherlock, SATO and $C^2$.}
     \label{fig:C2C_sherlock_sato}
\end{figure}

For evaluating Sherlock and SATO, we utilized their respective pre-trained models provided by them. Figure~\ref{fig:C2C_sherlock_sato} compares the accuracy of the top result from $C^2$, Sherlock and SATO\footnote{Their API did not provide subsequent result choices.}. The accuracy of $C^2$ is higher than both SATO and Sherlock for all datasets.  On an average, the accuracy of $C^2$ is $62.9\%$, Sherlock's is $26.6\%$ and SATO's is $26\%$. 
With respect to the datasets, $C^2$ provides maximum benefit over LIM and SEM datasets. This experiment  validates the effectiveness of $C^2$ to correctly identify concepts for columns containing textual, categorical and numerical values. $C^2$ and SATO consume about the same order of time while Sherlock takes about 6-7 times as much.

\subsection{Comparison with Colnet}
\begin{figure}[H]
\centering
\begin{subfigure}{.25\textwidth}
  \includegraphics[width=\columnwidth]{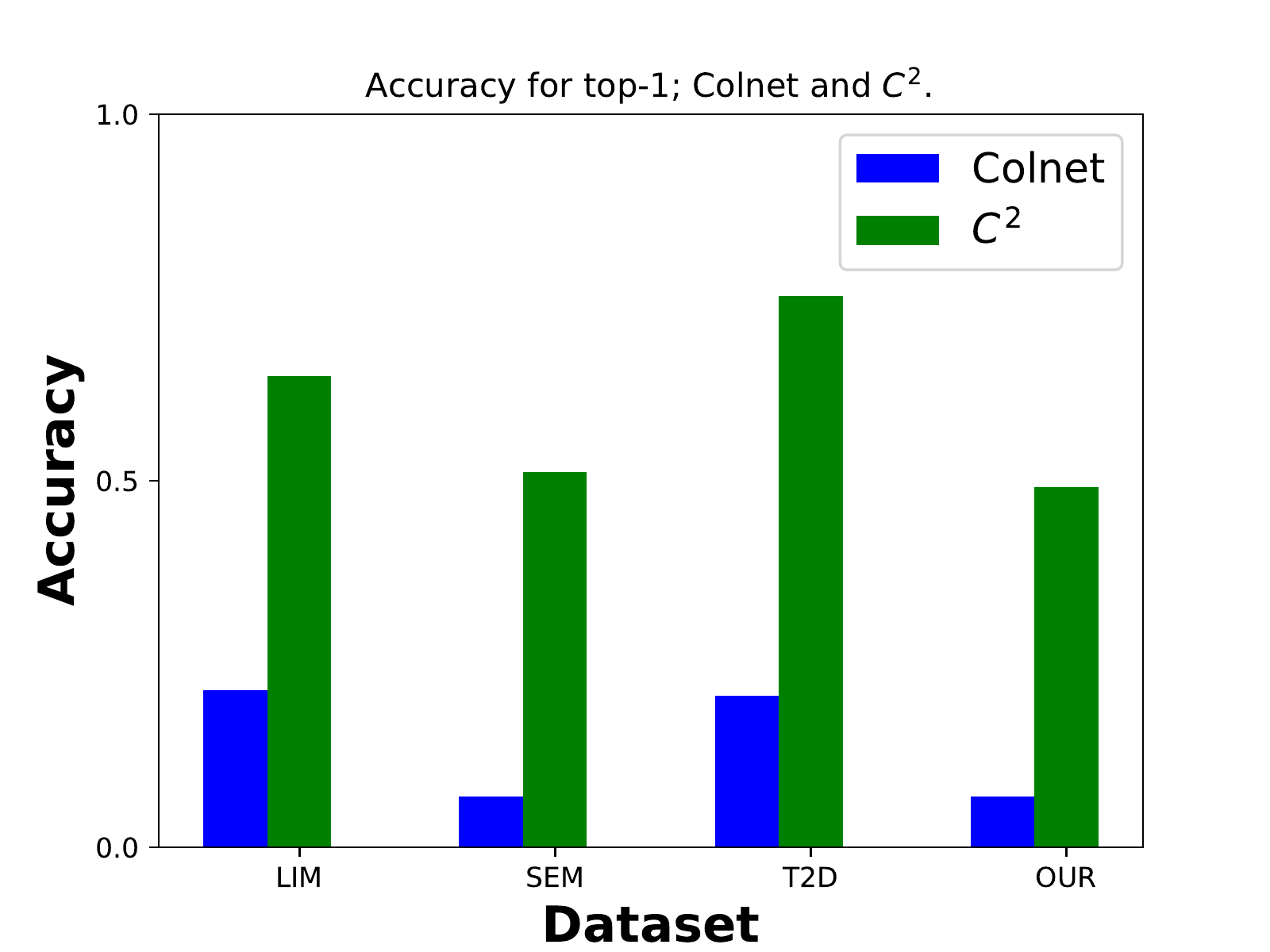}
  \caption{Accuracy for top result}
  \label{fig:sub1}
\end{subfigure}%
\begin{subfigure}{.25\textwidth}
  \includegraphics[width=\columnwidth]{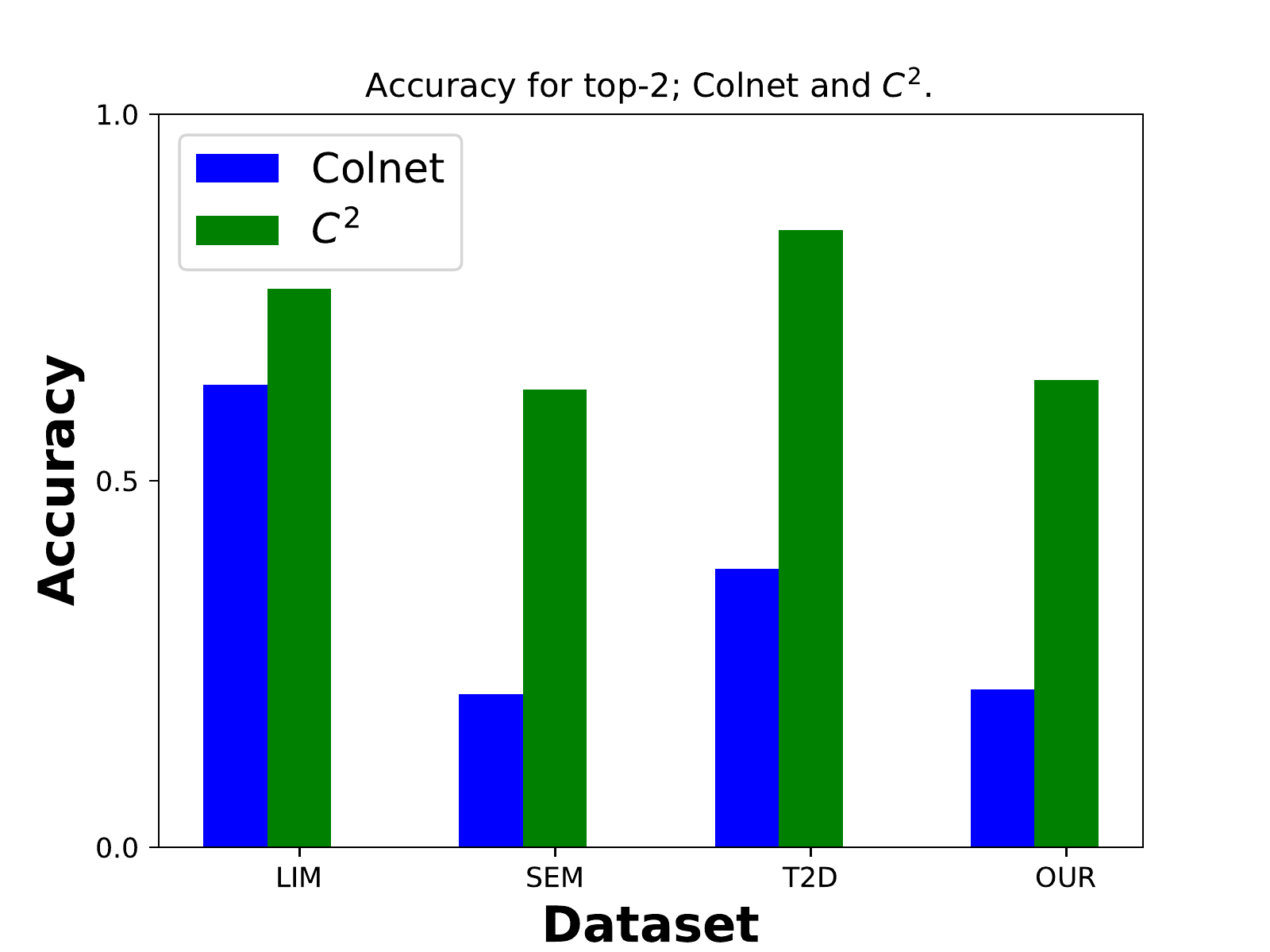}
  \caption{Accuracy within top-2}
  \label{fig:sub2}
\end{subfigure}
\caption{Comparing $C^2$ with Colnet (Top 1 and 2 results).}
\vspace{-5pt}
\label{fig:C2C_vs_colnet}
\end{figure}

Colnet does not provide a pretrained models and the training occurs afresh with respect to the entities provided in each test data and classifiers are then built specifically for classes in the test data. Training based on on each test dataset required more than 24 hrs for a dataset containing 150 concepts. It trains a different classifier for each concept, considers pair of cell values\footnote{This procedure generates quadratic number of pairs for each column. For efficiency reasons, we considered 200 values of each column to generate pairs of values.} and tests the input column with the classifier of all candidate concepts. As a result, the inference process takes a vast amount of time as well. As an example, it required more than $12$ hrs to identify the concepts for 411 columns in T2Dv2\footnote{This time includes the lookup time to search cell values over the KG.}.
We could only obtain results for 4 smaller datasets (LIM, SEM, T2Dv2, OUR) in a reasonable amount of time ($2$ days) per dataset. 
Figure~\ref{fig:C2C_vs_colnet} compares the quality of the top 1 and $2$ results. On an average for these 4 datasets, Colnet's top 1 and 2 results' accuracies are $13.9\%$ and $35.9\%$, respectively compared to $C^2$'s being $60\%$ and $71.6\%$, respectively.

HNN, which is closely related to Colnet, takes even significantly more time than Colnet, as it considers pairs across columns as well. None of the test data ran on it within 48 hours.


\section{Conclusion}
The approaches for annotating structured data so far have either relied on handcrafted rules or gold standard training data, both of which haven't demonstrated generalizability. 
Our method presented in this paper reflects a significant departure from both approaches. It is based on large, openly available, however somewhat noisy data collected from plethora of sources. 
The main contribution presented in this paper is an effective maximum likelihood approach through ensembles, and indirectly using ``wisdom of the crowds''. This far out-scales and out-generalizes both, the idea of hand-crafting each and every rule for pattern matching, or building a classifier for each concept upon collecting massive amounts of gold standard training data. Our next steps are to (a) investigate more advanced likelihood estimation methods; (b) incorporate better understanding of relationship between output concepts; (c) find ways to utilize supervised learning in the current framework.



\bibliographystyle{ACM-Reference-Format}
\bibliography{bib_c2c}


\typeout{get arXiv to do 4 passes: Label(s) may have changed. Rerun}

\end{document}